\documentclass[letterpaper, 10 pt, conference]{ieeeconf} 

\IEEEoverridecommandlockouts 
\usepackage{graphics} 
\usepackage{graphicx}
\usepackage{amsmath}
\usepackage{amssymb}
\usepackage{xspace}
\usepackage{algorithm} 
\usepackage{algorithmic} 
\usepackage{booktabs}

\usepackage{dsfont}
\usepackage{url}

\renewcommand{\S}{\mathcal{S}}

\newcommand{\A}{\mathcal{A}}

\newcommand{\algoName}{SEED\xspace}

\usepackage{xcolor}

\title{\LARGE \bf
Primitive Skill-based Robot Learning from Human Evaluative Feedback
}

\author{Ayano Hiranaka$^{* 1}$, Minjune Hwang$^{* 2}$, Sharon Lee$^{2}$, Chen Wang$^{2}$, Li Fei-Fei$^{2}$, Jiajun Wu$^{2}$, Ruohan Zhang$^{2}$
\thanks{*Equal contribution, alphabetically ordered}
\thanks{$^{1}$Department of Mechanical Engineering, $^{2}$Department of Computer Science, Stanford University, CA, USA, {\tt\small zharu@stanford.edu}}%
}

\begin{document}

\maketitle
\thispagestyle{empty}
\pagestyle{empty}

\begin{abstract}
Reinforcement learning (RL) algorithms face significant challenges when dealing with long-horizon robot manipulation tasks in real-world environments due to sample inefficiency and safety issues. To overcome these challenges, we propose a novel framework, \algoName, which leverages two approaches: reinforcement learning from human feedback (RLHF) and primitive skill-based reinforcement learning. Both approaches are particularly effective in addressing sparse reward issues and the complexities involved in long-horizon tasks. By combining them, \algoName reduces the human effort required in RLHF and increases safety in training robot manipulation with RL in real-world settings. Additionally, parameterized skills provide a clear view of the agent's high-level intentions, allowing humans to evaluate skill choices before they are executed. This feature makes the training process even safer and more efficient. To evaluate the performance of \algoName, we conducted extensive experiments on five manipulation tasks with varying levels of complexity. Our results show that \algoName significantly outperforms state-of-the-art RL algorithms in sample efficiency and safety. In addition, \algoName also exhibits a substantial reduction of human effort compared to other RLHF methods. Further details and video results can be found at \url{https://seediros23.github.io/}.
\end{abstract}

\section{Introduction}
Long-horizon manipulation tasks pose a significant challenge for robot learning due to the limitations of reinforcement learning (RL)  \cite{sutton2018reinforcement} in physical, real-world environments. While RL has shown remarkable success in simulation environments, its application to real-world robotics is hampered by sample inefficiency and safety concerns, as it is impractical to allow robots to engage in unbridled trial-and-error interactions with the physical environment for extended periods. Sparse reward signals in long-horizon tasks exacerbate these difficulties. In response, recent research has proposed two promising approaches to enhance RL in real-world robot applications: leveraging human evaluation and augmenting robots with primitive skills. Here, we present a novel framework that complements these approaches to tackle long-horizon manipulation tasks in the physical world.

First, different types of human guidance \cite{zhang2019leveraging} are often introduced to speed up learning and reduce risks. This is known as reinforcement learning from human feedback (RLHF). For instance, humans can provide real-time evaluative feedback (``good'', ``neutral'', or ``bad'') \cite{littman2015reinforcement,lin2020review} to indicate how desirable the observed behavior is. Evaluation is an attractive approach for robot learning because it is relatively easy to collect. For physical robot learning tasks, it may be infeasible for a human trainer to define a reward function (for RL) or provide a demonstration (as in imitation learning, IL) due to safety concerns or limited human expertise. Nevertheless, similarly to how sports coaches provide valuable feedback for professional athletes, it is still often possible for humans to guide the learning agent through useful evaluations. This underscores the potential utility of non-expert feedback in skill acquisition and mastery. Even in cases where RL or IL approaches are viable, evaluation can be used to increase the speed of task learning.

Another such approach is the augmentation of robots with a pre-defined library \cite{nasiriany2022augmenting} of parameterized primitive skills, such as \texttt{Pick(obj-A)} or \texttt{MoveTo(x,y)}. Although deep RL has the potential to learn a policy with low-level, high-dimensional action space like joint commands,  augmenting robots with skills has emerged as a promising approach to improve the efficiency and scalability of robot learning in physical environments.

\begin{figure}[t]
    \centering
    \includegraphics[width=1\linewidth]{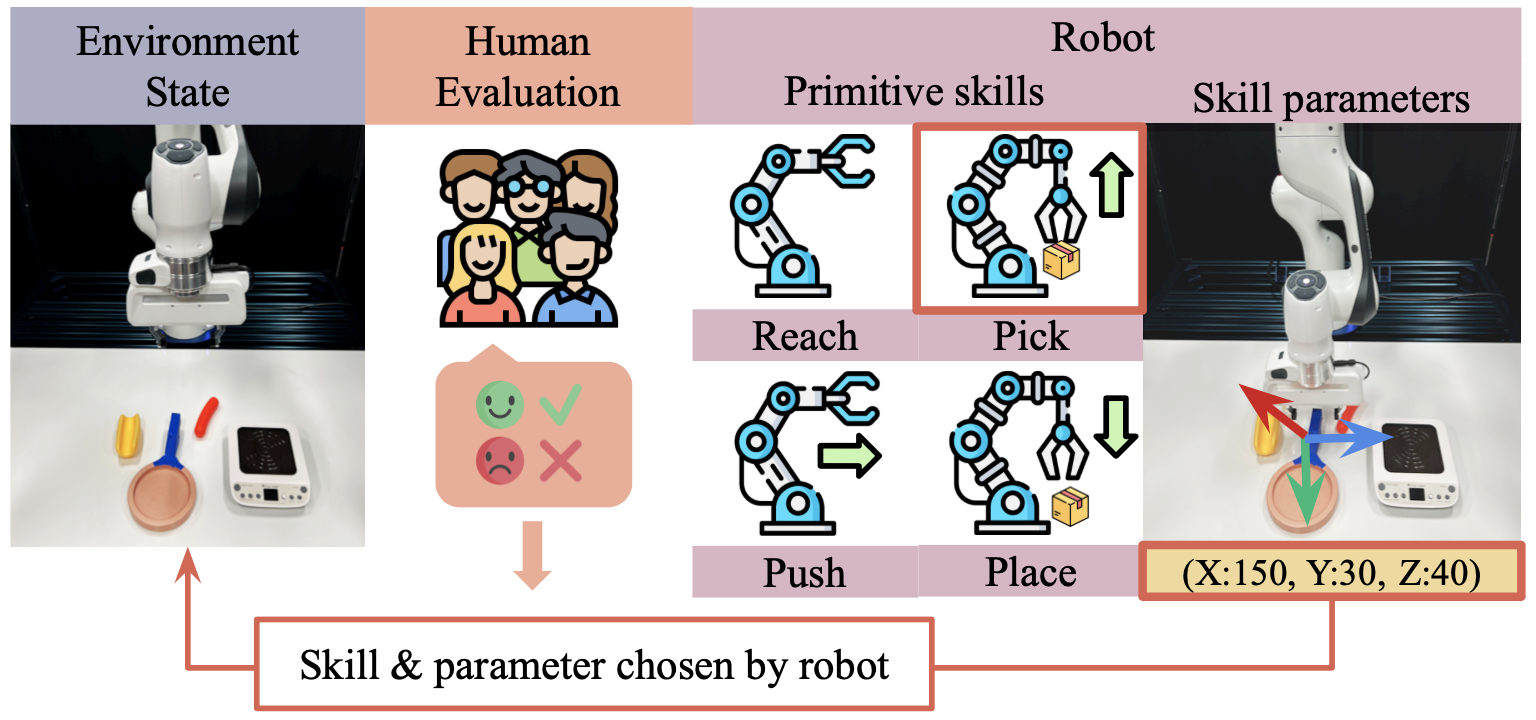}
    \caption{An overview of skill-based evaluative feedback (\algoName). Human trainers provide evaluative feedback on a robotic learning agent's choice of primitive skills and skill parameters. }
    \label{fig:pull}
\end{figure}

To overcome the challenges faced by robot learning in real-world manipulation tasks, we propose a novel framework, \algoName (\textbf{S}kill-based \textbf{E}valuative f\textbf{E}e\textbf{D}back), as shown in Fig.~\ref{fig:pull}, which synergistically integrates two approaches: learning from human evaluative feedback and primitive skill-based motion control. The combination of primitive skills and evaluative feedback is highly advantageous for RL agents for several reasons. Firstly, by breaking down complex, long-horizon tasks into a sequence of primitive skills, evaluative feedback can provide dense training signals, which makes long-horizon tasks with sparse rewards more tractable. Secondly, evaluating low-level robotic actions can be a time and resource-intensive task for humans~\cite{zhang2023dual}, but evaluating primitive skills require significantly less effort. Thirdly, primitive skills are intuitive and can reveal the high-level intentions of the robot, allowing humans to evaluate skill choices before they are executed. This ``evaluation without execution'' design is difficult for robots that only have low-level actions. The use of negative feedback from humans to prevent the robot from executing the action can ensure safety during RL training in real-world settings.

We conducted extensive experiments on five manipulation tasks of varying complexities in both the Robosuite \cite{zhu2020robosuite} simulator and in the real world. Our empirical results demonstrate that \algoName significantly outperforms alternative approaches in terms of sample efficiency, safety, and human effort, particularly in long-horizon tasks with sparse rewards. 

Our experiments also highlight emerging capabilities of \algoName, including zero-shot generalization ability in unseen scene configurations through reward composition. Additionally, \algoName outperforms imitation learning-based methods when suboptimal demonstrations or multimodal demonstrations are present. 

\section{Related Work}

\textbf{Learning from human evaluative feedback for tasks with sparse rewards.}
In this framework, human trainers monitor the learning process of an agent and provide a learning signal to indicate whether the observed behavior is desirable, in the form of continuous scalar signals \cite{knox2009interactively}, binary values ~\cite{thomaz2006reinforcement,najar2020interactively}, or trajectory-level critiques \cite{cui2018active} through different means of providing feedback~\cite{isbell2001social, tenorio2010dynamic,griffith2013policy,najar2020interactively,akinola2020accelerated}. The agent then learns a policy to maximize positive feedback from humans. Human evaluation is often interpreted as value function~\cite{knox2009interactively,warnell2018deep}, or advantage function~\cite{macglashan2017interactive,arumugam2019deep}. Human evaluation can be naturally combined with environment rewards so the agent learns simultaneously from both sources~\cite{knox2010combining,knox2012reinforcement,arakawa2018dqn}. Applying evaluative feedback-based RL to physical robots is challenging: \cite{zhang2023dual,knox2013training,wang2021apple,najar2016training} shows that this is feasible, but without primitive skills, we are limited to shorter-horizon tasks such as reaching and placing, or tasks with low-dimensional state and action space. 

\textbf{Leveraging primitive skills for long-horizon robot learning tasks.}
A plethora of recent research has explored leveraging parameterized skills to solve long-horizon robot manipulation tasks ~\cite{chitnis2022learning, nasiriany2022augmenting, zhu2021hierarchical, shridhar2022cliport, shridhar2023perceiver, xu2021deep, wang2022generalizable, cheng2022guided, agia2023stap, li2023behavior}. Traditional search-based algorithms, such as task and motion planning (TAMP) \cite{lozano2014constraint, toussaint2015logic, garrett2020pddlstream, garrett2021integrated}, have been widely utilized for effective multi-step parameterized skill optimization. However, these methods heavily depend on analytically-defined components, such as preimage functions and environment kinematics models. Recent learning-based approaches have been developed, leveraging deep neural networks to learn to solve long-horizon tasks from either human demonstrations~\cite{shridhar2022cliport, shridhar2023perceiver} or task rewards~\cite{xu2021deep, cheng2022guided, agia2023stap}. Although learning-based methods provide greater flexibility in solving complex tasks, they often require a significant number of demonstrations and well-defined reward functions for learning primitive skills which can be both costly and challenging to scale up. 

Our work is closely related to the approach introduced in MAPLE~\cite{nasiriany2022augmenting}, which aims to enhance the sample efficiency of learning manipulation policies by augmenting deep RL with parameterized skills. However, MAPLE still faces limitations that hinder its ability to generalize to novel scenes and imposes safety risks in deployment on real-world robots. In contrast, \algoName, addresses these issues by leveraging simple yet effective evaluative feedback as reward signals. Our approach not only significantly improves the sample efficiency of the model but also ensures the training process is safe and user-friendly in novel environments.

\section{Method}
Our method is designed to overcome challenges in learning long-horizon robot manipulation with deep RL. By leveraging RLHF and parameterized skills, we propose a novel framework --- \algoName, to improve sample efficiency, reduce human effort, and ensure safety in RL tasks with physical robots.

\begin{figure*}[t]
    \centering
    \includegraphics[width=0.9\linewidth]{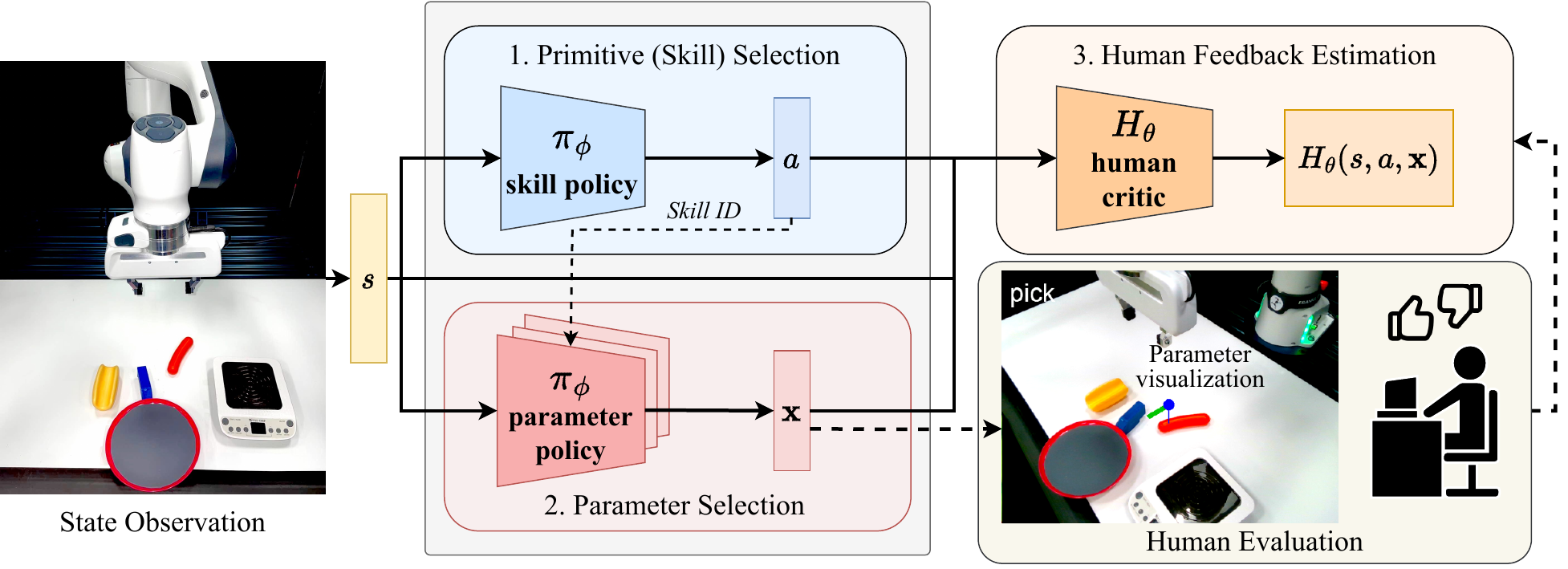}
    \caption{Neural network architecture for \algoName. The network consists of a critic network that predicts human evaluative feedback, a skill actor network that selects primitive skill, and a parameter actor network that selects parameters for the selected skill. Each skill has a unique parameter policy. The skill policy outputs the ID of the selected skill, which is used to invoke the parameter policy corresponding to the selected skill. Outputs of the skill and parameter networks are used by human trainers to provide evaluative feedback. This evaluation signal, combined with the skill and parameter selection, is used to train the human critic.}
    \label{fig:model}
\end{figure*}

\subsection{Parameterized Skills}
We represent the robot decision-making problem as a Markov decision process denoted by the tuple $\langle \S, \A, P, R, \gamma \rangle$, representing the state space, the action space, the transition function, the reward function, and the discount factor. A policy $\pi$ is a mapping from observation state space $\S$ to a probability distribution over the robot action space $\A$.

However, naively learning RL robot agents with low-level joint control or operational space control is impractical in real world due to the sample inefficiency and safety concerns. Since skill-augmented RL has shown promising results in solving long-horizon tasks with better sample efficiency, we follow recent works in learning RL agents with parameterized skills~\cite{nasiriany2022augmenting} and augment our action space $\A$ of the manipulation agent with the following primitive skills ($a$) and their parameters ($\mathbf{x}$):
\begin{itemize}
    \item \textit{Reaching:} Moves end-effector to location $(x,y,z)$.
    \item \textit{Picking:} Picks up an object at location $(x,y,z)$.
    \item \textit{Placing:} Places an object at location $(x,y,z)$.
    \item \textit{Pushing:} Reaches to starting location $(x,y,z)$ and pushes end-effector in $x$ or $y$ direction by $\delta$.
    \item \textit{Gripper Release:} Opens gripper (has no parameters).
\end{itemize}
By leveraging the parameterized skills, the control policy $\pi$ is able to focus on learning skill and parameter selection, which bypasses the burden of learning low-level motor control and improves learning efficiency. 

Since our decision-making algorithm does not require knowledge of the primitive skills' underlying control mechanism, the skills can be implemented in any method as long as they are robust and adaptive to various situations encountered during the task. In our implementation, each of the skills is predefined by closed-loop controllers that move the end effector in straight-line paths between a series of waypoints. Robosuite's built-in controller and Deoxys's API for Franka Emika Panda arm controller \cite{zhu2023viola} are used for the simulation and real-world experiments respectively. Operational space control (OSC) \cite{khatib1987unified} is used for both scenarios.

\subsection{Skill-based evaluative feedback.}
Leveraging human evaluative feedback can further improve sample efficiency and safety in long-horizon robot manipulation tasks. TAMER \cite{knox2009interactively, warnell2018deep} is a widely used framework for RL from evaluative feedback. Instead of using the environment reward, human trainers provide a scalar signal to indicate whether the observed decision is desirable or not. We denote this signal as $H(s,a,\mathbf{x})\in\{-1,0,+1\}$, where $s$ is the state vector, $a$ is the one-hot skill selection vector, and $\mathbf{x}$ is the skill parameter vector. Since part of the action space (skill parameters) is continuous, we use Soft Actor-Critic (SAC)~\cite{haarnoja2018soft} as the RL backbone. 
We use MAPLE~\cite{nasiriany2022augmenting} as the framework for simultaneously learning a skill-policy that selects a primitive skill, and unique parameter policies for each skill to select the skill parameters. The key difference is that MAPLE is purely based on RL, but \algoName's critic is trained using supervised learning where the objective is to predict human evaluative feedback. 

The model architecture is shown in Fig.~\ref{fig:model}. We can estimate human evaluative feedback $\hat{H}(s,a,\mathbf{x})$ using a critic head in SAC. We assume $\theta$, $\phi$, $\psi$,
parameterize the critic, the skill selection actor network, and the parameter selection actor network, respectively. The learning objective for the critic is an L2 loss:  
\begin{equation}
    \mathcal{L}(\theta) = \mathds{E}_{(s,a,\mathbf{x},H)\sim \mathcal{D}}\|\hat{H}_{\theta}(s,a,\mathbf{x}) - H(s,a,\mathbf{x}) \|^2_2
\label{eq:h_loss}
\end{equation}
Similar to MAPLE, we have separate loss functions for the skill actor and the parameter actor:
\begin{equation}
    \mathcal{L}(\phi) =
    \mathds{E}_{a \sim \pi_{\phi}} \Big[\alpha_{\phi} \log(\pi_{\phi}(a|s)) -  \mathds{E}_{\mathbf{x} \sim \pi_{\psi}}\hat{H}_{\theta}(s,a,\mathbf{x}) \Big] 
\label{eq:a_loss}
\end{equation}
\begin{equation}
    \mathcal{L}(\psi) =
    \mathds{E}_{a \sim \pi_{\phi}} \mathds{E}_{\mathbf{x} \sim \pi_{\psi}} \Big[\alpha_{\psi} \log(\pi_{\psi}(\mathbf{x}|s,a)) -  \hat{H}_{\theta}(s,a,\mathbf{x}) \Big], 
\label{eq:x_loss}
\end{equation}
$\alpha_{\phi},\alpha_{\psi}$ are the temperature parameter for the maximum entropy objective in SAC \cite{haarnoja2018soft,haarnoja2018soft2}. The actors update the policy distribution in the direction suggested by the critic. The agent learns a policy to maximize expected feedback from humans.  

\subsection{Balanced replay buffer.} During the initial stages of training with human feedback, the policy network mostly proposes suboptimal actions. As a result, the model's replay buffer will predominantly be filled with actions labeled with negative human feedback. Following prior works on resampling methods for imbalanced learning \cite{chawla2002smote, lane2012decision}, we sample an equal number of ``good" and ``bad" samples in each batch during off-policy learning stages, promoting faster convergence of the critic and the actor networks. In the absence of positive samples in the early training stage, we resort to the standard batch sampling approach using negative samples. This method can be considered to be a special case of prioritized replay buffer \cite{schaul2016prioritized}, in which we prioritize sampling transitions with positive human feedback.

\subsection{Facilitating learning with affordances.} MAPLE has shown that adding an affordance score as a small auxiliary reward can facilitate exploration and learning \cite{nasiriany2022augmenting}, e.g., a pushing skill is only appropriate in the vicinity of pushable objects, and the agent should be penalized with a negative reward for using the skill inappropriately. MAPLE utilizes well-crafted, skill-specific affordance scores that scale with distance from keypoints to encourage the agent to specify position parameters near important sites for each primitive. To accelerate learning in real robot experiments, we adopt a simplified version of MAPLE's affordance score where we add a small penalty of $-0.1$ when the skill parameter is not near any task-relevant objects. This affordance reward design is more general and involves less human engineering effort.

\subsection{Evaluation without execution.}  
Primitive skills and their parameters like those defined in MAPLE \cite{nasiriany2022augmenting} have clear semantics and are intuitive to humans. Therefore humans can evaluate robot's selection of skills and parameters even before the robots execute the action.

\begin{figure}
    \centering
    \includegraphics[width=0.9\linewidth]{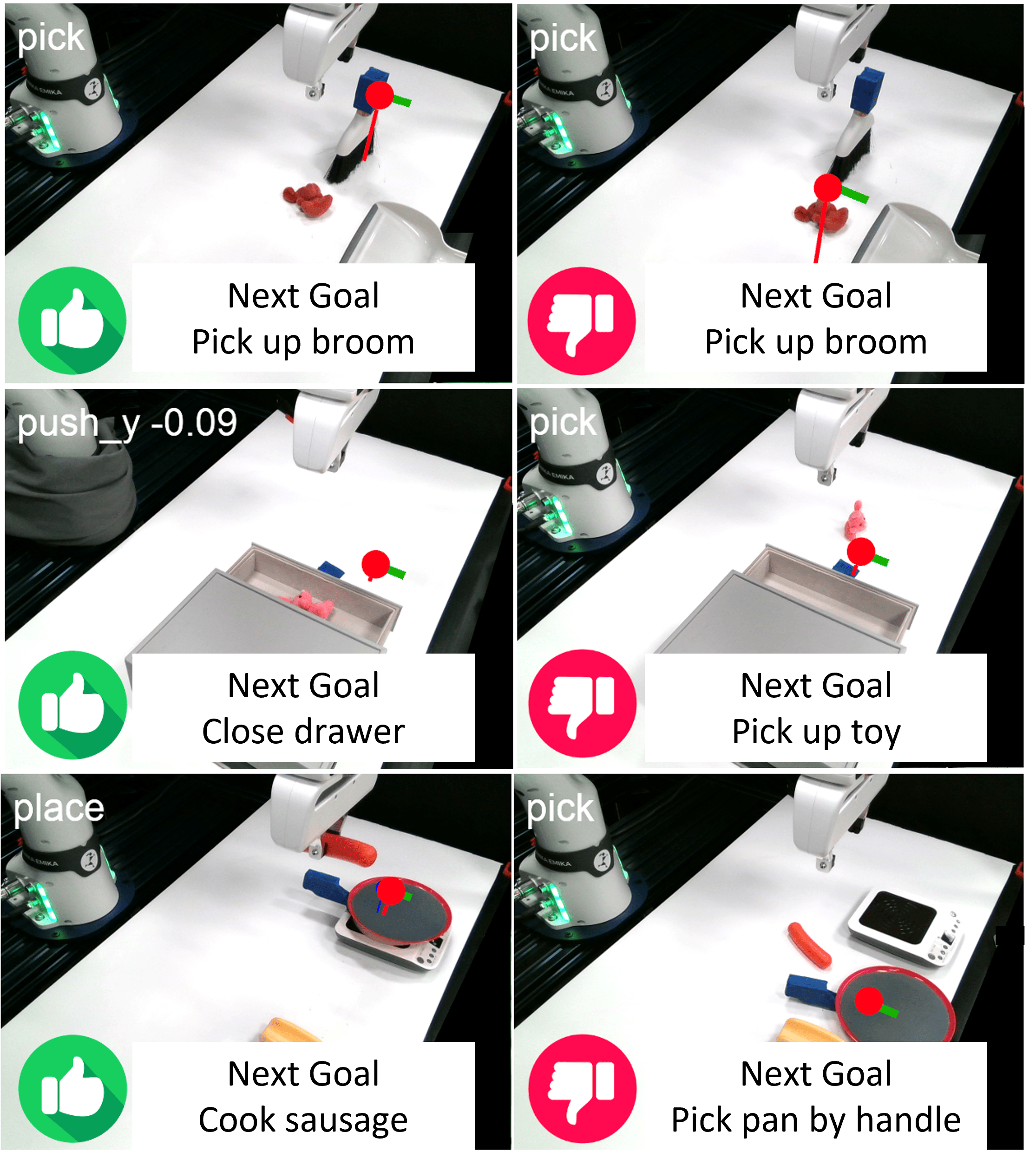}
    \caption{SEED user interface. The red dot and line show $(x,y,z)$ positions of the chosen skill parameter projected on the 2D camera view. Green bar shows the gripper orientation. For example, in the top right image, the next subgoal is to pick up the broom. Human gives negative feedback because the proposed parameters are too far away from the broom handle.}
    \label{fig:eval}
\end{figure}

The interface for evaluation without execution is shown in Fig.~\ref{fig:eval}. During the training process, a depiction of the robot's workspace, marked with annotations of the agent's skill and parameter selections, is presented to a human evaluator. During the initial stage of training, the human trainer evaluates each action as ``good'' or ``bad'' based solely on the visual representation of the robots' skills and parameters. Only when the human is confident of the robot's ability to make good decisions, will the robot be allowed to execute the action, making training more safe and efficient. The full pipeline of \algoName is shown in Algorithm 1.

\begin{algorithm}[t]
\label{alg:seed}
\caption{Skill-based Evaluative Feedback (\algoName)}
\begin{algorithmic}[1]
\STATE Initialize network weights for skill policy $\pi_{\phi}(a|s)$, parameter policy $\pi_{\psi}(\mathbf{x}|s,a)$, and human feedback critic network $H_{\theta}(s,a,\mathbf{x})$; initialize replay buffer $\mathcal{D}$
\FOR{episode = 1, \dots, N}
\STATE Initialize $t \leftarrow 0$
\WHILE{episode not terminated}
\STATE Sample skill $a_t = \pi_{\phi}(\cdot|s_t)$
\STATE Sample skill parameter $\mathbf{x}_t = \pi_{\psi}(\cdot|s_t, a_t)$
\STATE Query human for evaluative feedback $H_t(s_t,a_t, \mathbf{x}_t)$
\STATE Store transition ($s_t$, $a_t$, $\mathbf{x}_t$, $H_t$) in $\mathcal{D}$
\IF {$H_t(s_t,a_t,\mathbf{x}_t) = +1$}
\STATE Execute ($a_t, \mathbf{x}_t$) 
\ENDIF
\STATE Sample a minibatch of ($s_t$, $a_t$, $\mathbf{x}_t$, $H_t$) from $\mathcal{D}$ to perform gradient updates on $\theta, \phi, \psi$ based on Eqs~\ref{eq:h_loss}, \ref{eq:a_loss}, and \ref{eq:x_loss}. 
\STATE $t \leftarrow t+1$
\ENDWHILE
\ENDFOR
\end{algorithmic}
\end{algorithm}

\section{Experiments Setups}
\subsection{Baselines}
To understand the effect of human evaluation and to compare skill-based learning with low-level action-based learning, we compare \algoName with the following baselines:
\begin{itemize}
    \item \textbf{SAC} \cite{haarnoja2018soft} is the standard actor-critic algorithm that optimizes the stochastic policy with entropy regularization. 
    \item \textbf{TAMER} \cite{knox2009interactively} is an existing framework for  RLHF. To adapt TAMER to continuous actions space, we used TAMER+SAC, which replaces the standard critic with a human feedback critic which estimates a scalar signal from human trainers. Human trainer evaluates every low-level step, or a single command to the OSC controller. The agent is trained on dense human rewards and sparse environment rewards.
    \item \textbf{MAPLE} \cite{nasiriany2022augmenting} is an existing framework for RL with behavior primitives. Compared to SAC, MAPLE replaces the standard actor with a hierarchical policy that has a high-level policy that determines the skill selection, and a low-level parameter policy that determines the parameter selection given the primitive skill. This algorithm is trained on sparse environment rewards and does not involve human evaluations.
    \item \textbf{MAPLE-aff} is a variant of MAPLE that leverage affordance score as a dense reward signal \cite{nasiriany2022augmenting}. We intend to show that human feedback is a more powerful learning signal than this hand-designed affordance reward. This algorithm has no human evaluations.
\end{itemize}
TAMER and MAPLE can be viewed as ablated versions of \algoName. The former does not have primitive skills and the latter does not have human feedback. Training hyperparameters are shown in Table \ref{tab:hyperparams}. \textit{Train frequency} refers to the number of environment steps between each gradient descent, in which we take \textit{gradient steps} times of gradient updates.

\textbf{Synthetic human feedback in simulation.} We utilized synthetic feedback in simulated environments instead of real human feedback as in real-world environments. Synthetic feedback assumes idealized human feedback behaviors, which allows us to focus on the learning algorithms themselves and perform more extensive and controlled experiments. For \algoName, we utilize predefined heuristics (i.e., skill-specific affordance reward) to generate binary human feedback for each high-level step.

In contrast, TAMER was trained with an oracle which is a fully trained SAC agent, since heuristics for low-level actions are difficult to specify. Specifically, the TAMER agent chooses an action $a$ in state $s$, and the oracle chooses an action $a^*$. The oracle SAC computes the Q values for these actions: $Q(s,a)$ and $Q(s,a^*)$. If the learning agent chooses an action that has a Q-value close enough to $Q(s,a^*)$, it is a good action and the agent should receive positive feedback. Otherwise, it should receive negative feedback:
$  H(s,a) = +1, \text{if \ } Q(s,a) \geq \alpha Q(s,a^*) \ \text{and} -1 \ \text{otherwise}$.
The $\alpha$ value is initially set as $0.999$, and it increases over time to encourage the agent to learn to choose better actions during training. 

\textbf{Human feedback in real world}. For each of the three tasks, a single human trains the agent twice by providing evaluation signals via a keyboard key press.

\begin{figure*}[t]
    \centering
    \includegraphics[width=0.99\linewidth]{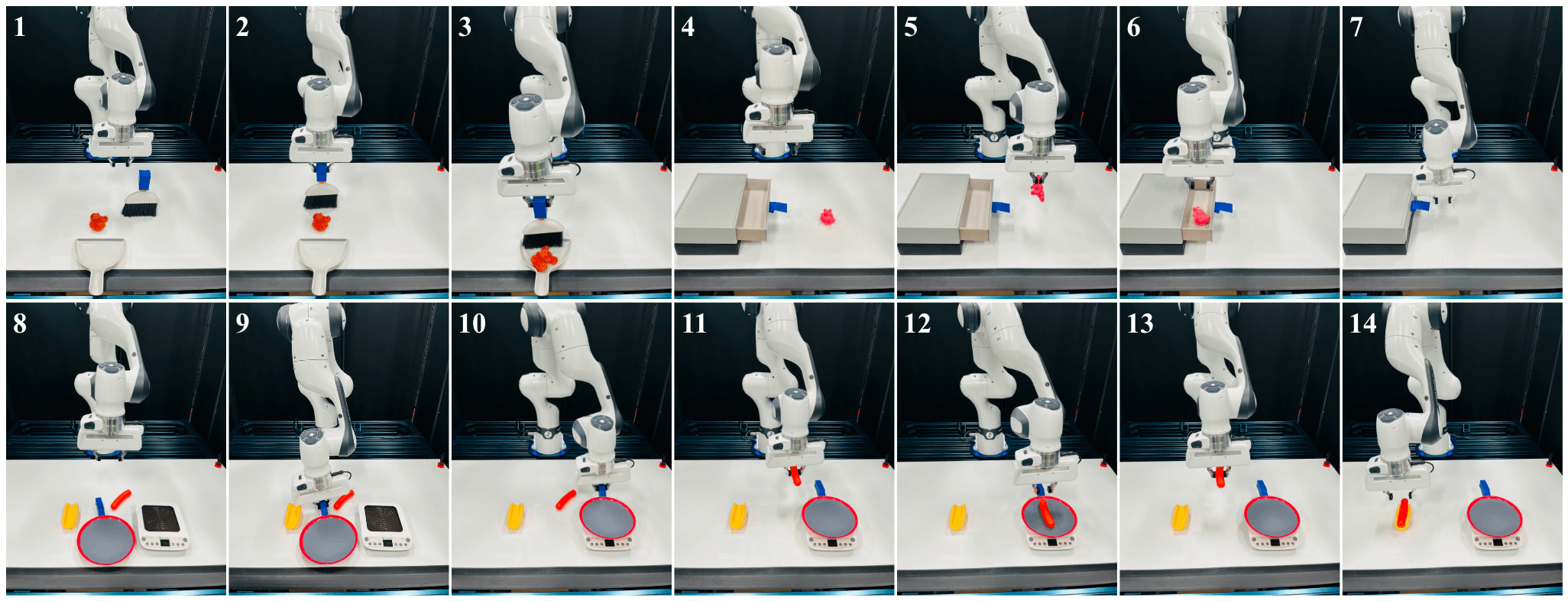}
    \caption{Visualizations of real-world, long-horizon manipulation tasks with intermediate steps.
    Top row: \texttt{Sweeping} (1-3) and \texttt{Collecting-Toy} (4-7) tasks; bottom row (8-14): \texttt {Cooking-Hotdog}, the task with the longest horizon.} 
    \label{fig:tasks}
\end{figure*}
\subsection{Long-horizon manipulation tasks}
The robot we use is a Franka Emika robot arm. The simulation tasks are implemented in the Robosuite~\cite{zhu2020robosuite}. We evaluate \algoName and baseline algorithms in simulation and in the real world, in the following long-horizon tasks (as shown in Fig.~\ref{fig:tasks}). Because our skill implementation is independent of robot proprioception, this information is omitted for \algoName and MAPLE, enabling reduced state space.

\texttt{Reaching} is a simulation task in which the robot has to move its gripper to the fixed area from a random starting location. The state, $s\in\mathbb{R}^4$ represents the gripper position in 3D and a binary state indicating the gripper status for all algorithms. In this environment, we provide \textit{Reaching} and \textit{Gripper Release} as available skills to skill-based algorithms; however, the desired outcome is for the model to learn to solely rely on the reaching primitive. This task is relatively short-horizon, and is mainly a sanity check for our implementation of baseline algorithms since many of them have zero performance in more challenging tasks described below.

\texttt{Stacking} is a simulation task in which the robot has to stack a small block on top of a larger block. The initial locations of both blocks and the robot are randomized. For low-level baselines, the state space $s\in\mathbb{R}^{10}$ comprises 3D positions of the gripper and blocks, as well as a binary state indicating whether the gripper is closed. On the other hand, MAPLE and \algoName utilize a reduced state space $s\in\mathbb{R}^6$, which includes 3D positions for both blocks. In this task, \textit{Picking} and \textit{Placing} are available. 

\texttt{Sweeping} is in the real world. The robot is required to pick up a broom and sweep a toy into a dustpan. For SAC and TAMER, the state space ($s\in\mathbb{R}^{10}$) includes the 3D positions of the gripper and broom, 2D position of the toy, the gripper state, and a flag indicating whether the broom is being grasped. MAPLE and \algoName's state space omits the gripper position and state. Available primitive skills for this task include \textit{Picking} and \textit{Pushing}.

\texttt{Collecting-Toy} is in the real world. The robot is tasked with picking up a toy, placing it in a drawer, and pushing the drawer closed. For SAC and TAMER, the state space ($s\in\mathbb{R}^{10}$) includes the 3D positions of the gripper and toy, the gripper state, the delta value of the drawer's current position from the closed position, and flags indicating whether the toy is being grasped and whether it is in the drawer. MAPLE and \algoName omit the gripper position and state from their state space. Available primitive skills for this task include \textit{Picking}, \textit{Pushing}, and \textit{Placing}.

\texttt{Cooking-Hotdog} is in the real world. The task requires the robot to perform a series of actions, including picking up a skillet and placing it on a stove, placing a sausage on the skillet, picking up the sausage again, and placing it in a bun. In both SAC and TAMER, the state space ($s\in\mathbb{R}^{15}$) contains 3D positions of the gripper, sausage, and skillet, as well as flags indicating the gripper state and the status of various steps in the task. However, the state space for MAPLE and \algoName does not include the gripper position or state. Available skills for this task are \textit{Picking} and \textit{Placing}.

\begin{table}[]
    \centering
    \caption{Training Hyperparameters (Sim / Real)}
    \begin{tabular}{c|c|c|c|c}
        & SAC & TAMER & MAPLE & SEED \\
        \hline
        learning rate & 3e-5 / - & 3e-5 / 3e-4 & 3e-3 / 3e-3 & 3e-3 / 3e-3 \\
        batch size  & 256 / - & 256 / 1024 & 256 / 1024 & 256 / 1024 \\
        $\gamma$ (discount rate) & 0.99 / - & 0.99 / 0.99 & 0.99 / 0.4 & 0.99 / 0.4 \\
        gradient steps & 5 / - & 5 / 30 & 5 / 3 & 5 / 30 \\
        train frequency & 1 / - & 1 / 25 & 1 / 2 & 1 / 25 \\
    \end{tabular}
    \label{tab:hyperparams}
\end{table}

\subsection{Hardware setup}
The robot we use is a Franka Emika robot arm. The experiment is run on a PC operating on Ubuntu 20.04 with \textsc{Intel}$^{\circledR}$ Core i7-7700K CPU and \textsc{Nvidia}$^{\circledR}$ GTX 1080 Ti graphics card. For object position estimation, two calibrated \textsc{Intel}$^{\circledR}$ Realsense\texttrademark Depth Camera D415 are used.  

\section{Results}
\textbf{Evaluation metrics.} The primary performance metric utilized in our simulations is the task success rate, as the algorithms can be trained until convergence. To measure the task success rate, we conducted 100 evaluations throughout the training process, each consisting of 10 rollouts. 

In real-world scenarios, extensive evaluations prove to be costly. As a result, we have adopted an approach wherein we measure the number of successes over the course of the training steps. This allows us to monitor the progress of the algorithms in real-time without incurring significant expenses. Furthermore, it is crucial to consider safety concerns while evaluating the performance of robots. We have identified two critical safety scenarios that need to be monitored: a safety violation resulting in damage to the robot or objects, and a safety violation leading to task failure. In the former scenario, the emergency stop button is pressed; in the latter case, a manual reset is required to restore normalcy. The count of safety violations for each of the trials is documented.

\begin{figure}
    \centering
    \includegraphics[width=0.9\linewidth]{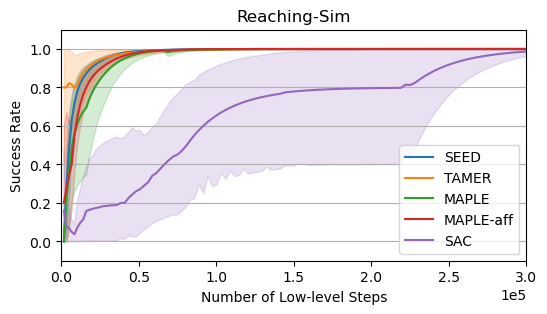}
    \includegraphics[width=0.9\linewidth]{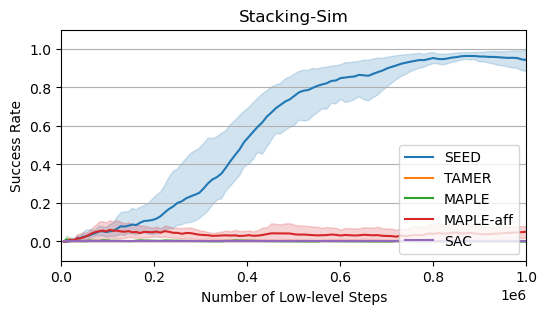}
    \caption{Average success rate over training steps (low-level steps, one low-level action per step) for \texttt{Reaching} and \texttt{Stacking} in simulation. \algoName learns to solve the tasks more efficiently compared to the baselines. Error bars indicate the standard error of the means ($n=5$).}
    \label{fig:sim_result}
\end{figure}

\textbf{\algoName is sample efficient.}
Simulation experiment results are shown in Fig.~\ref{fig:sim_result} comparing the performance of several RL algorithms on two different robotic manipulation tasks: \texttt{Reaching} and \texttt{Stacking}. In the simple, short-horizon task, \texttt{Reaching}, both TAMER and \algoName algorithms exhibit rapid learning which highlights the advantage of using evaluative feedback. As expected, MAPLE and MAPLE-aff algorithms demonstrate faster learning rates than the SAC algorithm in this task. However, in the more complex and challenging \texttt{Stacking} task, \algoName outperforms all other algorithms by a substantial margin. It is worth noting that although MAPLE-aff may eventually learn the \texttt{Stacking} task after four million steps, as reported in the original research \cite{nasiriany2022augmenting}, \algoName learns to solve the task in only 800,000 steps.

While the original work on MAPLE-aff demonstrated the advantage of skill-based actions in sim2real transfer \cite{nasiriany2022augmenting}, this approach is only applicable when a digital twin setting can be prepared. However, this is not always possible, calling for methods that allow the robot to train from scratch in the real world.

\begin{figure}
    \centering
    \includegraphics[width=0.9\linewidth]{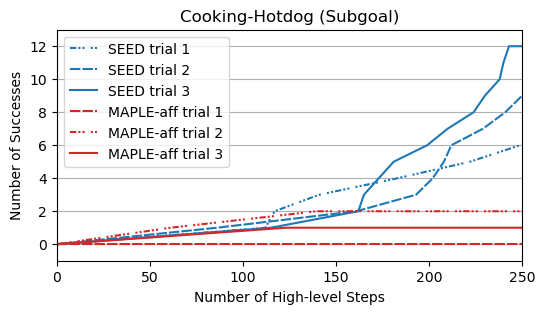}
    \caption{Number of successes over training steps (high-level steps, one skill per step) for the first goal in \texttt{Cooking-Hotdog} (picking up the skillet). \algoName learns to solve this subgoal more efficiently than MAPLE-aff. }
    \label{fig:real_result1}
\end{figure}

The first challenge in training MAPLE-aff from scratch in real-world settings is the time-consuming process. To overcome this challenge, the \algoName algorithm employs evaluation without execution and relies on human feedback to optimize training time, which is around ten times faster than MAPLE-aff. The second challenge of training MAPLE-aff is that physical robots pose a safety risk. Nonetheless, to compare the performance of \algoName and MAPLE-aff, we conducted experiments using a simplified version of the \texttt{Cooking-Hotdog} task, which involves only the first subgoal of picking up the skillet. The results shown in Fig.~\ref{fig:real_result1}, indicate that on average, \algoName can successfully complete the subgoal nine times within 250 high-level steps, while MAPLE-aff only succeeds once. Given the low success rate of MAPLE-aff and safety concerns associated with continuing the experiments, we did not run MAPLE-aff on the entire task.

\textbf{\algoName ensures better safety.}
Table~\ref{tab:safety} presents the safety violation ratio, which is given by the number of safety violations divided by the total number of decision steps. Our analysis reveals that \algoName exhibits significantly lower safety risks when compared to MAPLE-aff. The chance of safety violation in MAPLE-aff is about 3 to 7 times compared to \algoName. The main reason is that \algoName enables evaluation without execution, which can prevent dangerous actions from being executed. The findings of our study indicate that \algoName holds great promise in enhancing safety in robot learning, a crucial consideration for real-world applications.

\begin{figure}
    \centering
    \includegraphics[width=0.85\linewidth]{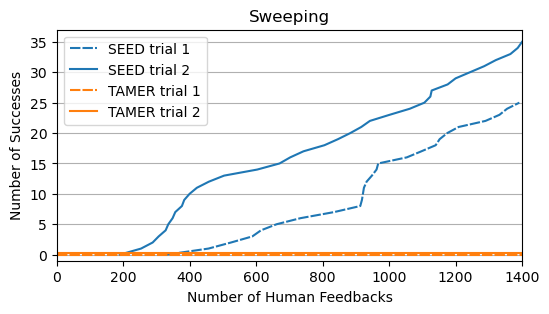}
    \includegraphics[width=0.85\linewidth]{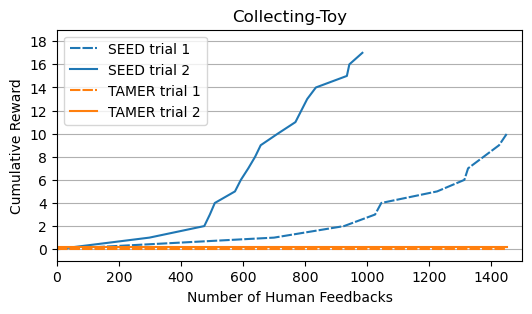}
    \includegraphics[width=0.85\linewidth]{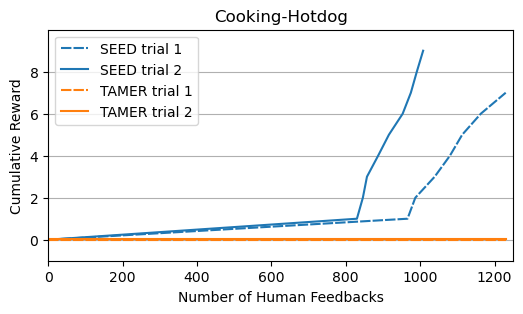}
    \caption{Number of successes over the number of human feedback for three real-world tasks. \algoName learns to solve all tasks efficiently while TAMER cannot. The experiments are terminated when the maximum number of steps is reached or the agent has learned to solve the task consistently.}
    \label{fig:real_result2}
\end{figure}

It is worth noting that in the case of TAMER, one decision step is equivalent to one low-level step, whereas in MAPLE-aff and \algoName, one decision step corresponds to one primitive skill step, which typically involves around 100 low-level steps. Therefore the risk of TAMER is underestimated here (and its performance is zero as shown in Fig.~\ref{fig:real_result2}).

\begin{table}[]
    \centering
    \caption{Safety violation ratio.}
    \begin{tabular}{c|ccc}
    \toprule    
    & TAMER & MAPLE-aff & \algoName \\ 
    \midrule
    \texttt{Sweeping}  & 0.25\% &  8.50\% &  1.11\% \\
    \texttt{Collecting-Toys} & 0.26\%  & 1.19\%  & 0.40\% \\
    \texttt{Cooking-Hotdog} & 0.49\% & 3.54\% & 0.51\% \\        
    \bottomrule
    \end{tabular}
    \label{tab:safety}
\end{table}

\textbf{\algoName significantly reduces human effort.}
Due to concerns over the MAPLE-aff algorithm's sample efficiency and potential safety risks in a physical robot setting, we opted to exclude it from the rest of real-world experiments. Instead, we focused solely on training the TAMER and \algoName algorithms until task completion. We compare \algoName and TAMER based on the amount of human effort required in real-world experiments. Figure~\ref{fig:real_result2} displays the results obtained from providing both TAMER and \algoName with the same quantity of human feedback. Notably, \algoName was able to learn effectively within the given amount of feedback, while TAMER failed to achieve any successful task completion.
Remarkably, for all three long-horizon tasks, \algoName has successfully learned to solve them. Additionally, human trainers adapt quickly and learn how to provide better feedback for the robots, as evidenced by the much better results observed in the second trial compared to the first. Please refer to the supplemental video for a detailed analysis of the learning results.

\section{Conclusion}
Real-world manipulation tasks that involve long horizons present numerous challenges to robotic learning agents, including safety guarantee and sample efficiency. In addition, if human data are required, as in the case of RLHF, it is essential to minimize the associated human effort. This work presents \algoName, an innovative approach that synergistically integrates human evaluative feedback and primitive skills to enhance the efficiency and safety of real-world reinforcement learning. The proposed method overcomes the challenges associated with real-world long-horizon manipulation tasks, thereby paving the way for future research to scale up robot learning with improved safety guarantees and affordable human costs.

\section*{Acknowledgments}
This work was in part supported by ONR MURI N00014-22-1-2740, ONR MURI N00014-21-1-2801, the Stanford Institute for Human-Centered AI (HAI), Amazon, Analog Devices, JPMC, and Salesforce. Ruohan Zhang is partially supported by Wu Tsai Human Performance Alliance Fellowship.

\bibliography{ref}
\end{document}